\documentclass[runningheads]{llncs}

\usepackage{eccv}

\usepackage{eccvabbrv}
\usepackage{multirow} 
\usepackage{graphicx}
\usepackage{booktabs}
\definecolor{mygreen}{RGB}{50,205,50}
\usepackage[accsupp]{axessibility}  

\usepackage{hyperref}
\usepackage{bbm}
\usepackage{tabularx}
\usepackage{array}
\usepackage{orcidlink}

\begin{document}

\title{Uncertainty-Weighted Fusion of Image and Synthetic Event for Video Anomaly Detection} 

\titlerunning{Uncertainty-Weighted Fusion of Image and Synthetic Event}

\author{Sungheon Jeong\inst{1}\orcidlink{0000-0003-3540-7065} \and
Jihong Park\inst{2} \and
Mohsen Imani\inst{1}\orcidlink{0000-0002-5761-0622}}

\authorrunning{S.~Jeong et al.}

\institute{University of California, Irvine, CA 92697, USA \and
Independent Researcher}

\maketitle
\begin{abstract} 
    Most existing video anomaly detectors rely on RGB frames alone, which limit their ability to capture abrupt or transient motion cues that are critical for identifying anomalous events. We propose Uncertainty Weighted Image Event Fusion (IEF-VAD), a framework that integrates complementary RGB and synthetic motion information through a principled weighting mechanism. The method models the high variance and heavy tailed characteristics of synthetic motion cues with a Student’s t likelihood, computes value level inverse variance weights using a Laplace approximation to prevent the image modality from overshadowing motion information, and performs iterative refinement to suppress residual cross modal noise. This formulation provides a more balanced and reliable fusion process compared to cross attention or gating based approaches that often suffer from modality dominance. Without requiring an event camera or frame level annotations, IEF-VAD achieves new state of the art performance on multiple real world anomaly detection benchmarks and remains stable under degradation applied to individual modalities. The results indicate that extracting and integrating complementary motion cues is an effective direction for robust video understanding across diverse environments.
\end{abstract}

\section{Introduction}
    Recent advances in deep learning have accelerated progress in multimodal data analysis~\cite{xu2023multimodal, liang2024survey, girdhar2023imagebind, radford2021learning, li2024mini, guo2019deep, akbari2021vatt}, enabling models to integrate diverse sensory inputs for complex tasks such as video anomaly detection~\cite{flaborea2023multimodal, zhang2024holmes, tang2024hawk, ji2020multi, feng2023self}. Prior work in this field has primarily emphasized static image information that provides strong spatial context. Dynamic motion cues, which encode subtle temporal changes, abrupt transitions, and transient patterns, remain comparatively underutilized. Real event data offers microsecond temporal resolution and polarity information suitable for capturing fleeting anomalies that RGB frames often fail to detect~\cite{gallego2020event, rebecq2019events, wang2022exploiting, chakravarthi2024recent}. However, the scarcity of large scale event datasets and sensor dependent variability has restricted their adoption in anomaly detection. To address this limitation, we construct synthetic motion cues from RGB videos using simple inter frame differencing and thresholding. While these cues do not replicate the temporal precision or polarity characteristics of true event streams, they retain the essential property of events by producing sparse and motion focused representations. We treat them as event like proxies that complement RGB information and can be effectively utilized within our uncertainty aware fusion framework.
    
    The challenge of integrating homogeneous data sources such as static RGB information and its derived synthetic motion cues under uncertainty remains unresolved. Recent transformer based fusion approaches for video anomaly detection~\cite{feng2021convolutional, wu2020not, zhou2019anomalynet, zhu2013variational, dinglearnable, ghadiya2024cross} continue to prioritize the detailed spatial structure of images, which causes motion representations to be under utilized despite their relevance to transient anomalies. Because both streams originate from a single visual modality, their homogeneity makes naive fusion especially vulnerable to modality dominance, where dense spatial features overwhelm motion focused information~\cite{park2024assessing, tsai2019multimodal, liu2021towards}. These issues motivate a fusion strategy that can explicitly consider uncertainty and rebalance contributions across homogeneous but imbalanced feature spaces.
    
    To address these challenges, we introduce an uncertainty aware framework that fuses RGB features with synthetic motion cues through a principled precision based mechanism. Our uncertainty weighted Image and Event proxy Fusion framework for Video Anomaly Detection (\emph{IEF-VAD}) estimates the reliability of each modality and assigns inverse variance weights derived from Bayesian principles. IEF-VAD reduces the influence of less reliable signals and prevents the spatially rich RGB stream from dominating the temporally informative motion cues. Heavy tailed noise in motion representations is captured with a Student's $t$ likelihood and approximated with a Gaussian via a Laplace expansion~\cite{malmstrom2023fusion, zhu2013variational, daxberger2021laplace, wu2021student}, aligning the fusion process with established Bayesian uncertainty estimation techniques~\cite{zhu2013variational, subedar2018uncertainty, kendall2017uncertainties, ober2021promises, gal2016dropout}. This formulation yields a balanced fused representation that preserves motion driven signals even in RGB dominant settings.
    
    Building on this formulation, we further refine the fusion at the value level by assigning each latent dimension its own variance estimate. Each feature is modelled with a Student's $t$ likelihood to characterize value specific heavy tailed noise and approximated using a Laplace based Gaussian expansion. The resulting per dimension variances are converted into precision weights for value level inverse variance fusion, ensuring that unreliable features are down weighted while motion related cues retain influence in the presence of strong RGB content.
    
    Furthermore, the framework includes a sequential update mechanism followed by an iterative refinement process. The sequential update combines each new image and motion cue observation with the previously fused state through inverse variance weights that mirror the Kalman gain in the classical filter formulation~\cite{welch1995introduction, haarnoja2016backprop}. The iterative refinement stage then focuses on residual errors that the fusion step cannot fully resolve, including feature level inconsistencies, modality specific noise, and small scale imbalances that arise when two streams have different noise characteristics. By repeatedly estimating and removing these residuals, the refinement network progressively denoises and rebalances the fused representation, resulting in a cohesive latent state.
    
    \noindent Our contributions are as follows:
    \begin{itemize}
        \item 
        \textbf{Precision Weighted Fusion Mechanism:} We present an uncertainty aware fusion strategy that assigns inverse variance precision weights to image and motion cue features, drawing on Bayesian inference and Kalman filtering. This mechanism adaptively balances contributions by reducing the influence of unreliable signals and prevents RGB features from overwhelming motion focused cues.
    
        \item 
        \textbf{Practical Integration of Synthetic Motion Cues:} We show that motion cues derived from standard RGB videos can be effectively incorporated into anomaly detection models. This provides a scalable alternative to real event sensors and extends motion aware fusion to a wider range of video datasets.
    
        \item 
        \textbf{Empirical Validation on Real World Datasets:} The proposed framework attains state of the art results on four benchmarks: UCF Crime (88.67\% AUC), XD Violence (87.63\% AP), ShanghaiTech (97.98\% AUC), and MSAD (92.90\% AUC). Beyond overall performance, our analysis indicates that motion cues play a critical role in motion dominant anomaly categories such as \emph{Assault}, \emph{Fighting}, and \emph{Shoplifting}, where they yield substantial gains over RGB only and conventional fusion baselines.
    \end{itemize}
    
\section{Related Work}
    \textbf{Adaptive Multimodal Fusion.}
        Multimodal fusion seeks to combine complementary signals from vision, audio, and language~\cite{xu2023multimodal, zong2023self, huang2022multi}. Earlier approaches relied on direct feature concatenation~\cite{ngiam2011multimodal}, while recent methods leverage transformers~\cite{tsai2019multimodal, akbari2021vatt, girdhar2022omnivore, li2021align} and contrastive pretraining~\cite{radford2021learning, girdhar2023imagebind, dai2022one} to improve cross modal alignment. Attention based architectures offer flexible integration but still suffer from modality dominance~\cite{liang2024survey, xu2023multimodal}. Another line of work employs large language models to modulate modality usage dynamically~\cite{shen2023multimodal, zhao2023mm, gong2023multimodal, li2023blip2, driess2023palme}, while others reweight modalities under estimated corruption for robustness~\cite{almarri2025robusta}. In contrast to these strategies, we highlight the role of explicit uncertainty as a control signal for stable fusion under imbalance and noise.

    \noindent\textbf{Event Cues for Video Anomaly Detection.}
        Event based sensors provide sparse and high temporal resolution signals~\cite{gallego2020event, zheng2023deep, chakravarthi2024recent}, enabling progress in recognition~\cite{shiba2022secrets, yang2023event, luo2023learning} and modality alignment with CLIP~\cite{wu2023eventclip, zhou2024eventbind, jeong2024expanding}. Video anomaly detection, in contrast, has primarily advanced through MIL based weak supervision~\cite{sultani2018real, wu2020not, zhong2022gtad, chen2024prompt} and remains largely image centric~\cite{liu2021weakly, georgescu2021anomaly}. Recent studies incorporate additional cues such as text through large language models~\cite{alayrac2022flamingo, driess2023palme, zhao2023mm}. We take a complementary direction by leveraging \emph{synthetic motion cues}, event like proxies derived directly from RGB videos~\cite{rebecq2019events, zhu2018ev, astrid2021synthetic, jeong2024expanding}, to capture motion salient dynamics without requiring dedicated event sensors.
    
    \noindent\textbf{Uncertainty and Bayesian Fusion.}
        Uncertainty estimation helps models adapt their confidence under noise and distributional shift~\cite{kendall2017uncertainties, gal2016dropout, ovadia2019can, subedar2018uncertainty}. Bayesian techniques such as dropout based inference and inverse variance weighting~\cite{gal2016dropout, subedar2018uncertainty, daxberger2021laplace} improve robustness in multimodal settings. We build on this line by combining a Laplace approximation with a Student's $t$ noise model~\cite{malmstrom2023fusion, wu2021student}, enabling principled fusion of image and motion cue features under heavy tailed uncertainty.
\section{Uncertainty-Weighted Fusion}
    \label{method}
    \begin{figure*}[ht]
        \centering
        \vspace{-5mm}
        \includegraphics[width=1.0\linewidth]{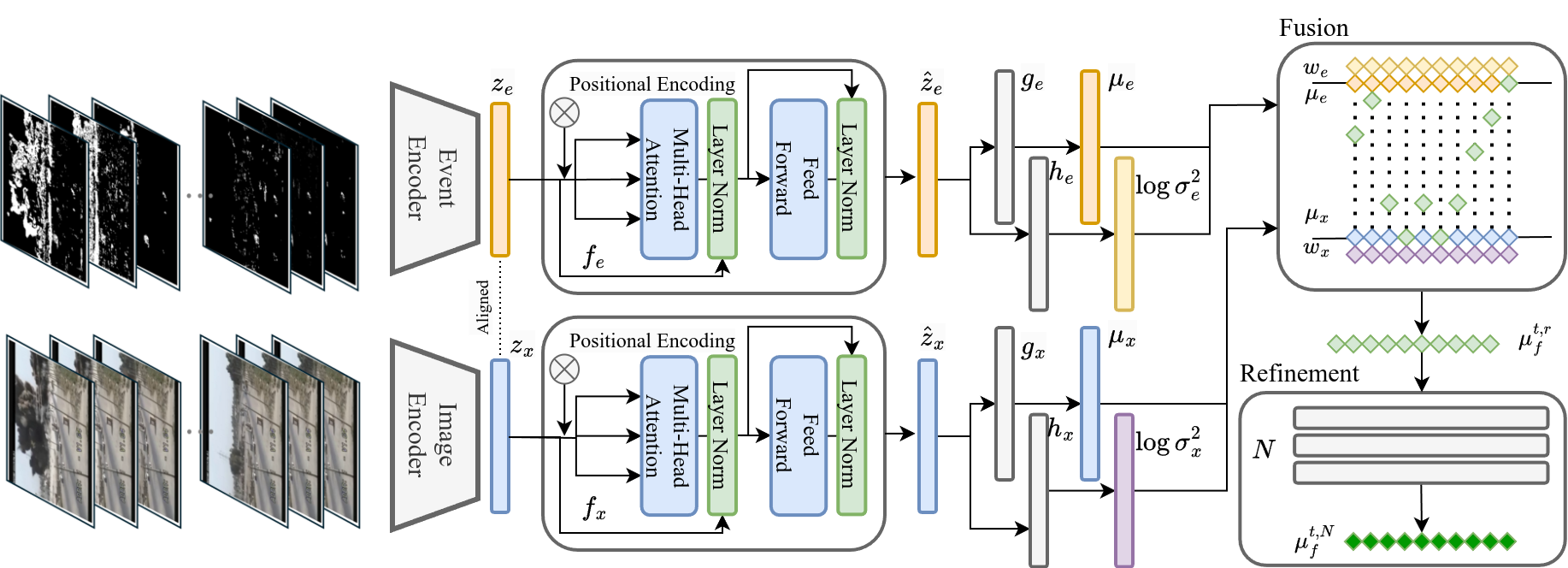}
        \caption{
            \textbf{Overview of IEF-VAD framework.} Each video frame and its corresponding synthetic motion cue are processed by CLIP encoders to obtain feature embeddings $z_m$. These are further encoded by modality-specific transformers $f_m$ to produce $\hat{z}_m$, which are then passed through projection heads $g_m$ and $h_m$ to estimate $\mu_m$ and $\sigma_m$. The estimated $\sigma_m$ is used to compute the uncertainty-aware fusion weight $w_m$, which is applied to obtain the initial fused representation $\mu_f^{t, 0}$. This representation is refined over $N$ iterative steps through a refinement network to produce the final output $\mu_f^{t, N}$.
        }
        \label{fig1}
    \end{figure*}

    In this work, as illustrated in Fig.~\ref{fig1}, we extract image embeddings ($z_x$) and motion embeddings ($z_e$) separately from videos, where $z_e$ denotes an \emph{event-like proxy} derived from RGB frames. Both streams observe the same scene and share a common spatial structure while capturing complementary and modality specific cues. $z_x$ encodes appearance related attributes such as color, background, and spatial detail, whereas $z_e$ focuses on transient changes and temporal dynamics. To integrate these homogeneous yet imbalanced representations, we introduce a fusion strategy that combines Bayesian uncertainty estimation with Kalman style update principles, where uncertainty reflects the reliability of each value level feature. We estimate latent features and their associated uncertainties for each stream and fuse them using a refinement network that iteratively corrects residual noise, misalignment, and modality specific artifacts, producing a stable fused representation $\mu_f$.

    \begin{figure*}[ht]
        \centering
        \includegraphics[width=1.0\linewidth]{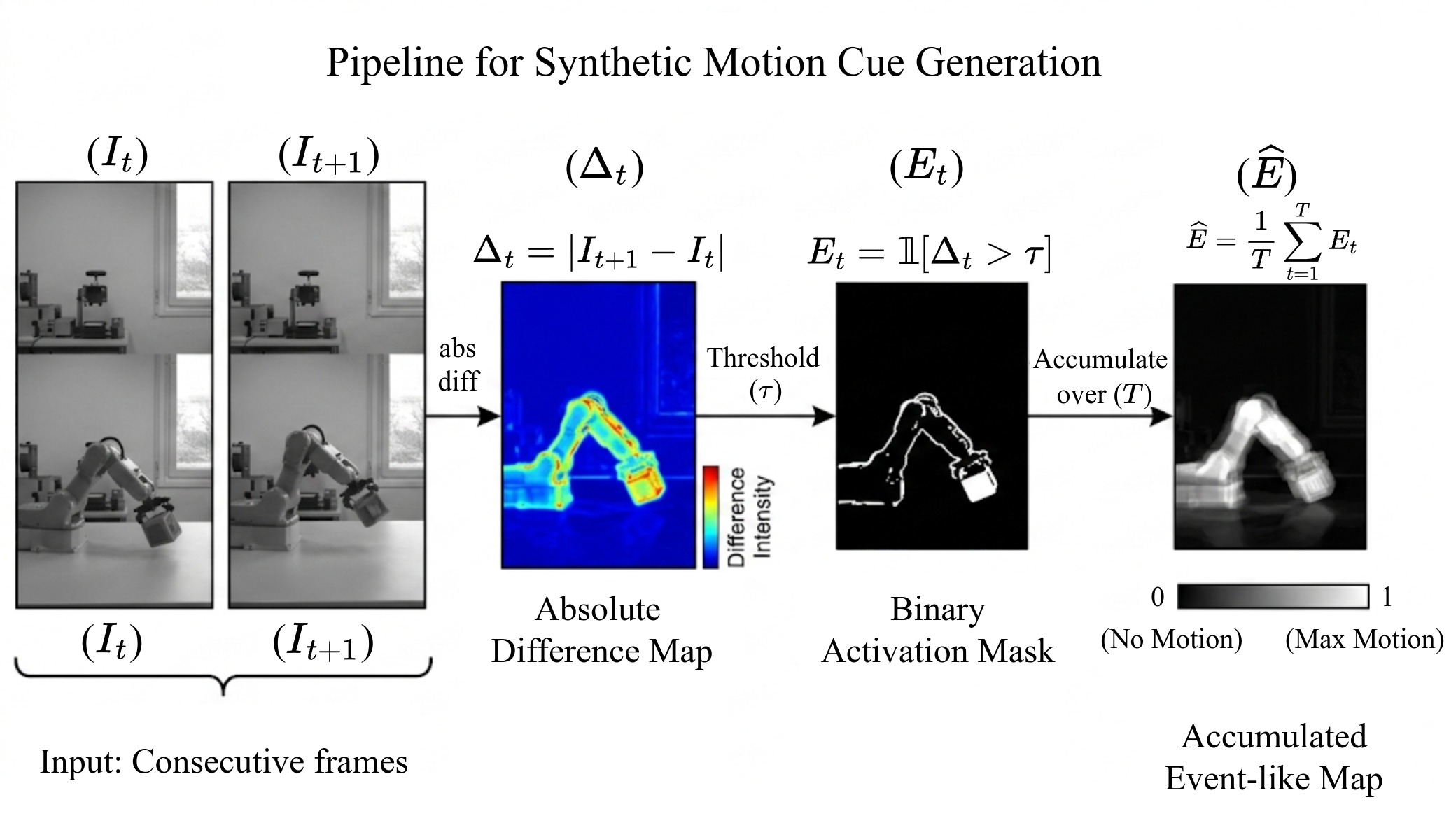}
        \caption{
            From consecutive grayscale frames $I_t$ and $I_{t+1}$, we compute $\Delta_t=\lvert I_{t+1}-I_t\rvert$, threshold it to obtain $E_t=\mathbf{1}\{\Delta_t>\tau\}$, and average activations over $T$ frames to form the event-like map $\hat{E}=\frac{1}{T}\sum_{t=1}^{T}E_t$.
        }
        \label{fig0}
    \end{figure*}

    \subsection{Synthetic Motion Cue}
        We derive a \textit{synthetic motion cue} directly from RGB frames to capture short term temporal dynamics. Given consecutive grayscale frames $I_t$ and $I_{t+1}$, we compute the absolute difference
            \begin{equation}
                \Delta_t(x,y) = \big| I_{t+1}(x,y) - I_t(x,y) \big|.
            \end{equation}
        Pixels are activated when the difference exceeds a threshold, $E_t(x,y)=\mathbbm{1}\;\! [\Delta_t(x,y) $ $> \tau ]$, where $\mathbbm{1}[\cdot]$ returns $1$ if the condition holds and $0$ otherwise, and the activations are accumulated over a temporal window to form the final motion cue Fig.~\ref{fig0}:
            \begin{equation}
                \hat{E}(x,y) = \frac{1}{T}\sum_{t=1}^{T} E_t(x,y).
            \end{equation}
        This event like representation suppresses static background and isolates localized motion, providing a lightweight proxy for event data without requiring specialized sensors. Although derived from RGB, it complements appearance features by emphasizing transient motion cues that are often diluted in image based representations, and consistently improves anomaly detection performance.

    \subsection{Modality Design and Uncertainty Estimation}
        We treat each modality as a noisy observation of the same scene:
        \begin{equation}
            z_m = \mu_m + \delta_m,\quad 
            \delta_m \sim t_\nu(0, \Sigma_m), \quad 
            m \in \{x, e\}.
        \end{equation}
        Here, $\mu_m\in\mathbb{R}^{d}$ denotes the latent estimate and $\Sigma_m=\operatorname{diag}(\sigma_m^{2})$ represents per-dimension variability. The Student’s $t$ model is chosen for its heavy-tailed behavior, which offers robustness when one modality becomes unreliable due to outliers or degradation. The degrees of freedom $\nu$ controls the tail heaviness: small $\nu$ yields heavier tails that down-weight large residuals more aggressively, while as $\nu\to\infty$ the distribution approaches a Gaussian.
        
        Given embeddings $z_m\in\mathbb{R}^{B\times T\times D}$, where $B$ is the batch size, $T$ the segment length, and $D{=}768$ the CLIP ViT-L/14 embedding dimension, we transpose them to $(T,B,D)$, process them with a modality-specific transformer $f_m$, and transpose back:
        \begin{equation}
            \hat{z}_m = (f_m(z_m^\top))^\top.
        \end{equation}
        
        Layer normalization stabilizes scale differences across modalities, and two projection heads estimate the posterior mean and log-variance:
        \begin{equation}
            \label{eq:extract}
            \mu_m = g_m(\hat{z}_m),\qquad
            \log\sigma_m^{2} = h_m(\hat{z}_m).
        \end{equation}
        The predicted variance $\sigma_m^{2}$ serves as the modality specific uncertainty, indicating the reliability of each latent dimension and determining its influence during the subsequent fusion steps. Under this formulation, $z_m$ can be interpreted as a sample from the posterior $t_\nu(\mu_m, \Sigma_m)$, where $\mu_m$ determines the modality’s contribution in fusion and $\sigma_m^{2}$ regulates its influence in the update stages.
        
        For efficiency, we adopt diagonal covariance matrices, which avoid the $\mathcal{O}(D^{3})$ complexity of full covariance estimation while enabling value-level variance modeling essential for balanced fusion across the two homogeneous modalities.
        
    \subsection{Inverse Variance Calculation}
        To determine the confidence of each modality, we compute inverse variance weights. Since direct variance manipulation under the Student’s $t$ likelihood is intractable, we approximate the distribution with a Gaussian using a Laplace expansion around its mode. The model predicts log-variances via Eq.~\eqref{eq:extract}, which yield $\sigma_m^2 = \exp(\log\sigma_m^2)$ under a Gaussian model. For the Student's $t$ case, we begin from its log-density
        \begin{equation}
            \label{eq:log_var}
            \log p(\delta)
            = -\frac{\nu+1}{2}\log\left(1+\frac{\delta^2}{\nu\sigma^2}\right) + C.
        \end{equation}
        
        Near the mode $\delta=0$, the Laplace approximation replaces the density with a Gaussian by matching second-order curvature. Using $\log(1+x)\approx x$ for small $x$, the quadratic term becomes proportional to $\delta^2/(\nu\sigma^2)$, leading to an effective Gaussian variance
        $
            \tilde{\sigma}^2 = {\nu}/({\nu+1})\;\sigma^2,
        $
        whose logarithmic form is
        \begin{equation}
            \label{eq: log_var2}
            \log\tilde{\sigma}^2
            = \log\sigma^2 + \log\!\left(\frac{\nu}{\nu+1}\right).
        \end{equation}
        The effective variance $\tilde{\sigma}_m^{2}$ serves as the modality-specific uncertainty, which directly governs its precision weight in the fusion process. This scaling reflects the heavier tails of the Student’s $t$ distribution and yields a conservative estimate of modality reliability.
        
        The resulting effective variance $\tilde{\sigma}_m^2$ is used to compute precision weights:
        \begin{equation}
            \label{eq: uncertainty_weights}
            w_m = \frac{1}{\tilde{\sigma}_m^2 + \epsilon},
        \end{equation}
        where $\epsilon$ ensures numerical stability. Modalities with smaller variance receive larger precision, meaning their estimates contribute more strongly during fusion. This follows standard Bayesian weighting principles and ensures that the fusion mechanism allocates influence according to each modality’s reliability, even under heavy-tailed noise.
        
    \subsection{Uncertainty-Weighted Fusion}
        We compute the fused representation $\mu_f \in \mathbb{R}^{B\times T\times D}$ by weighting the modality specific means using the precision scores defined in Eq.~\eqref{eq: uncertainty_weights}. The fused estimate is written as $\mu_f = \frac{w_x \mu_x + w_e \mu_e}{w_x + w_e}$. For the Student's t model, the precision scores are obtained from the effective variances produced by the Laplace approximation. Using Eq.~\eqref{eq: log_var2}, the effective variance is $\tilde{\sigma}_m^{2} = \exp\big(\log\sigma_m^{2} + \log(\frac{\nu}{\nu+1})\big)$, and its inverse determines the confidence of each modality. The precision $w_m$ is therefore the inverse of the modality-specific uncertainty encoded in $\tilde{\sigma}_m^{2}$, linking uncertainty directly to its contribution during fusion.
        
        Modalities with smaller variance or effective variance receive larger precision values, which increases their contribution in the fused representation. This formulation follows the standard Bayesian rule in which the posterior mean is a precision weighted combination of individual estimates. It also corresponds to the role of the Kalman gain in sequential estimation, where measurements with higher precision have a stronger influence on the updated state. This weighting mechanism ensures that informative motion cues are not overshadowed by the more spatially dense RGB stream.
        
    \subsection{Time Step Sequential Update}
        We extend the uncertainty weighted fusion to the temporal dimension by updating the fused state across time steps $T$ in a recursive manner, similar to Kalman style estimation but using the effective variances obtained from the Student's t model.
        
        If the predicted log variance at time $t$ is $\log\sigma_t^{2}$, the effective variance is
        \begin{equation}
            \tilde{\sigma}_t^{2} = \exp\left(\log\sigma_t^{2} + \log\left(\frac{\nu}{\nu+1}\right)\right)
            \label{eq:effective-variance}
        \end{equation}
        which corrects the predicted variance according to the heavy tailed noise. The effective variance $\tilde{\sigma}_t^{2}$ represents the uncertainty carried from each time step, allowing the model to propagate uncertainty through the temporal update.
        
        At the initial step $t=0$, we set the fused state directly from the first input as $\mu_f^0 = \mu^0$ and $\tilde{\sigma}_{f,0}^{2} = \tilde{\sigma}_0^{2}$. For subsequent steps $t \ge 1$, we fuse the previous state with the current input using inverse effective variance weights. The precision for the previous state is $w_f^{t-1} = 1/(\tilde{\sigma}_{f,t-1}^{2} + \epsilon)$ and the precision for the current input is $w^t = 1/(\tilde{\sigma}_t^{2} + \epsilon)$, where $\epsilon$ ensures numerical stability. The updated estimates are
        \begin{equation}
            \mu_f^{t} = \frac{w_f^{t-1}\mu_f^{t-1} + w^{t}\mu^{t}}{w_f^{t-1} + w^{t}}, \quad
            \tilde{\sigma}_{f,t}^{2} = \frac{1}{w_f^{t-1} + w^{t}}, \quad t \ge 1.
        \end{equation}
        
        This sequential procedure extends the static fusion by incorporating temporal continuity. Each update balances the previous state and the current observation according to their effective precisions, allowing the fused representation to accumulate information over time while remaining robust under the heavy tailed characteristics of the Student's t noise.

    \subsection{Iterative Refinement of Fused State}
        After the sequential fusion step, the fused representation may still contain small residual errors because each modality contributes its own noise characteristics and reliability estimate. Because the effective variances encode modality-specific uncertainty, the refinement network implicitly corrects the residual uncertainty that remains after the sequential fusion step. To further reduce these residuals, we employ an iterative residual-correction procedure, analogous to a Newton-style fixed-point update.
        
        Starting from the initial fused state $\mu_f^{0}$, we update the representation by repeatedly predicting and subtracting the residual error. At refinement step $r$ with $r=0,1,\dots,N$, a refinement network $F(\cdot)$ estimates the residual $\Delta\mu_f^{t,r} = F(\mu_f^{t,r})$ based solely on the current fused state. The update rule applies a step size $\lambda_r \in (0,1)$ that damps each correction:
        $\mu_f^{t,r+1} = \mu_f^{t,r} - \lambda_r \Delta\mu_f^{t,r}$.
        
        As iterations progress, the residuals are gradually reduced and the fused representation becomes more consistent with the underlying latent structure implied by both modalities. This refinement process acts as a controlled residual-correction mechanism that improves the quality of the fused features and produces a more stable representation for downstream tasks.
    
    \subsection{Video Anomaly Detection with Loss Functions}
        \textbf{Classification Loss ($\mathcal{L}_\text{cls}$):}
            After obtaining the refined fused representation $\mu_f^{t,N} \in \mathbb{R}^{B\times T\times D}$, we perform binary anomaly classification at each time step. A lightweight classification head $H$ maps $\mu_f^{t,N}$ to logits,
            $
                \hat{y} = H(\mu_f^{t,N}) \in \mathbb{R}^{B\times T\times 1},
            $
            followed by a sigmoid activation. The binary cross-entropy loss is computed against the frame-level labels:
            \begin{equation}
                \begin{aligned}
                \mathcal{L}_\text{cls}
                &= -\frac{1}{BT}
                \sum_{b=1}^{B}\sum_{t=1}^{T}
                \Big[
                y_{b,t}\log(\hat{y}_{b,t})
                + (1-y_{b,t})\log(1-\hat{y}_{b,t})
                \Big].
                \end{aligned}
            \end{equation}
            To handle variable temporal granularity and weak supervision without requiring frame-level annotations, we follow the segment aggregation strategy of~\cite{sultani2018real}. The sequence is divided into non-overlapping segments of 16 frames, and the predictions within each segment are averaged to obtain an instance-level score~\cite{ilse2018attention, sultani2018real}.
        
        \textbf{KL Divergence Loss ($\mathcal{L}_{\text{KL}}$):}
            Under the Student's t noise model, each modality contributes a latent variable with measurement noise $\delta_m \sim t_\nu(0,\sigma_m^{2})$. Because computing the KL divergence between a Student's t distribution and a standard normal prior is intractable, we use the effective variance $\tilde{\sigma}_m^{2}$ obtained from the Laplace approximation in Eq.~\ref{eq:effective-variance} and approximate the latent distribution as $\mathcal{N}(\mu_m,\tilde{\sigma}_m^{2})$. The KL term becomes
            \begin{equation}
                \begin{aligned}
                \mathcal{L}_{\text{KL}}
                &= KL\big(\mathcal{N}(\mu_m,\tilde{\sigma}_m^{2}) \| \mathcal{N}(0,I)\big) \\
                &= \frac{1}{2}\big(\tilde{\sigma}_m^{2} + \mu_m^{2} - 1 - \log\tilde{\sigma}_m^{2}\big).
                \end{aligned}
            \end{equation}
            This regularization constrains the latent representation using its uncertainty, preventing overconfident estimates when the Student's t noise is heavy-tailed.
            
            \textbf{Regularization Loss ($\mathcal{L}_\text{reg}$):}
            To align the modality-specific latent means, we introduce
            \begin{equation}
                \mathcal{L}_\text{reg}
                = \lambda_1 (1 - \cos(\mu_x,\mu_e))
                + \lambda_2 \big|\,\|\mu_x\| - \|\mu_e\|\,\big|,
            \end{equation}
            where $\lambda_1$ and $\lambda_2$ balance the angular and magnitude terms. This encourages semantic consistency between the two modalities and prevents scale imbalance that would otherwise distort the uncertainty-based fusion weights.
        
        \textbf{Overall Loss:}
            The final objective combines the classifier loss, the KL terms, and the alignment penalty:
            \begin{equation}
            \label{eq:loss}
            \mathcal{L}
            = \mathcal{L}_\text{cls}
            + \sum_{m}\mathcal{L}_\text{KL}^{m}
            + \mathcal{L}_\text{reg}.
            \end{equation}
            This composite loss ensures accurate anomaly predictions and stable latent representations, while the effective variances from the Laplace approximation keep all KL terms in closed form for efficient Gaussian-based optimization.

\section{Experiment Results}
    \label{experiment}
    We evaluate our method on four public video anomaly detection datasets, UCF-Crime~\cite{sultani2018real}, XD-Violence~\cite{wu2020not}, ShanghaiTech~\cite{liu2018future}, and MSAD~\cite{msad2024}, which span a wide range of real surveillance conditions. After applying standardized preprocessing—where synthetic motion cues are generated with threshold $\tau{=}10/255$, clamp value $10$, and temporal window $T{=}16$—and independently encoding each modality with its own transformer, we report performance using AUC and AP metrics. Additional analyses examine sensitivity to key hyperparameters such as the degree of freedom parameter, the number of samples, the refinement weight, and the stability constant, and also study the influence of different loss configurations. We further assess robustness under outlier injection and characterize the behavior of the weighting mechanism when the inputs are perturbed. Overall, the method combines complementary spatial and temporal information from images and motion cues in a weakly supervised framework, freezing the CLIP encoders and training only $f_m$, $g_m$, $h_m$, $F$, and $H$ with SGD (batch size $32$, $100$ epochs).

\begin{table*}[h]
    \centering
    \resizebox{\textwidth}{!}{%
    \begin{tabular}{lcccc|cc}
        \toprule
        \multirow{2}{*}{Method} & \multicolumn{2}{c}{UCF-Crime~\cite{sultani2018real}} & XD-Violence~\cite{wu2020not} & Shanghai-Tech~\cite{liu2018future}& \multirow{2}{*}{Method} & MSAD~\cite{msad2024} \\
        & AUC (\%) & Ano-AUC (\%) & AP (\%) & AUC (\%) &  & AUC (\%) \\
        \midrule
        Sultani et al. \cite{sultani2018real} & 84.14 & 63.29 & 75.18 & 91.72 & MIST (I3D) \cite{feng2021mist} & 86.65 \\
        Wu et al. \cite{wu2020not} & 84.57 & 62.21 & 80.00 & 95.24 & MIST (SwinT) \cite{feng2021mist} & 85.67 \\
        AVVD \cite{wu2022weakly} & 82.45 & 60.27 & - & - & UR-DMU \cite{zhou2023dual} & 85.02 \\
        RTFM \cite{tian2021weakly} & 85.66 & 63.86 & 78.27 & 97.21 & UR-DMU (SwinT) \cite{zhou2023dual} & 72.36 \\
        UR-DMU \cite{zhou2023dual} & 86.97 & 68.62 & 81.66 & 97.57 & MGFN (I3D) \cite{chen2023mgfn} & 84.96 \\
        UMIL \cite{lv2023unbiased} & 86.75 & 68.68 & - & 96.78 & MGFN (SwinT) \cite{chen2023mgfn} & 78.94 \\
        VadCLIP \cite{wu2024vadclip} & 88.02 & 70.23 & 84.51 & 97.49 & TEVAD (I3D) \cite{chen2023tevad} & 86.82 \\
        STPrompt \cite{wu2024weakly} & 88.08 & - & - & 97.81 & TEVAD (SwinT) \cite{chen2023tevad} & 83.6 \\
        OVVAD \cite{wu2024open} & 86.40 & - & 66.53 & 96.98 & EGO \cite{dinglearnable} & 87.36 \\
        \midrule
        Fusion (Cross Attention) & 86.57 & 69.03 & 84.62 & 97.36 & Fusion (Cross Attention) & 92.07  \\
        Fusion (Gating) & 87.86 & 69.74 & 84.62 & 97.71 & Fusion (Gating) & 92.75 \\
        \midrule
        IEF-VAD (Gaussian) & 88.11 $\pm$ 0.28 & 70.48 $\pm$ 0.66 & 87.18 $\pm$ 0.55 & 97.91 $\pm$ 0.08 & IEF-VAD (Gaussian) & 92.27 $\pm$ 0.34 \\
        IEF-VAD (Student-T) & \textbf{88.67} $\pm$ 0.45 & \textbf{71.50} $\pm$ 1.02 & \textbf{87.63} $\pm$ 0.54 & \textbf{97.98} $\pm$ 0.07 & IEF-VAD (Student-T) & \textbf{92.90} $\pm$ 0.27 \\
        \bottomrule
    \end{tabular}%
    }
    \caption{
        Comparison of various methods on multiple anomaly detection benchmarks, including UCF-Crime \cite{sultani2018real}, XD-Violence \cite{wu2020not}, ShanghaiTech \cite{liu2018future}, and MSAD \cite{msad2024}. All metrics are reported as the mean ($\pm$1 standard deviation) of 10 runs.
    }
    \label{tab:anomaly_comparison}
\end{table*}
    
    \subsection{Real-World Anomaly Detection in Surveillance Video}
        Table~\ref{tab:anomaly_comparison} shows that \emph{IEF-VAD} surpasses all prior weakly supervised detectors on every benchmark. On UCF-Crime, the Gaussian variant already matches the strongest published AUC of 88.11 percent, and the Student t extension further increases AUC to 88.67 percent while raising the anomaly focused AUC to 71.50 percent, which is an absolute improvement of about 1.3 percentage points over the previous best value of 70.23 percent reported by VadCLIP~\cite{wu2024vadclip}. The same pattern holds for XD-Violence with 87.63 AP and for ShanghaiTech with 97.98 AUC, where both variants outperform the strongest alternatives. On the recent MSAD benchmark, the Student t model obtains 92.90 AUC, improving upon the best I3D based baseline value of 86.82 AUC by more than six percentage points. Fusion baselines that rely on cross attention or gating tend to collapse toward a single dominant modality, while IEF-VAD maintains consistently higher performance. Reported standard deviation margins confirm that the improvements hold across ten independent runs.        
        
        These results confirm two key insights of our framework. First, value-level uncertainty weighting enables practical exploitation of the heterogeneous synthetic motion cue, turning its complementary signals into measurable performance gains whenever the RGB stream is degraded. Second, modelling heavy-tailed noise via a Student’s-\emph{t} likelihood yields further, systematic improvements, especially on long-tailed datasets (UCF-Crime, XD-Violence) where RGB frames frequently contain motion blur or illumination changes.

\begin{figure*}[h]
    \centering
    \includegraphics[width=1.0\linewidth]{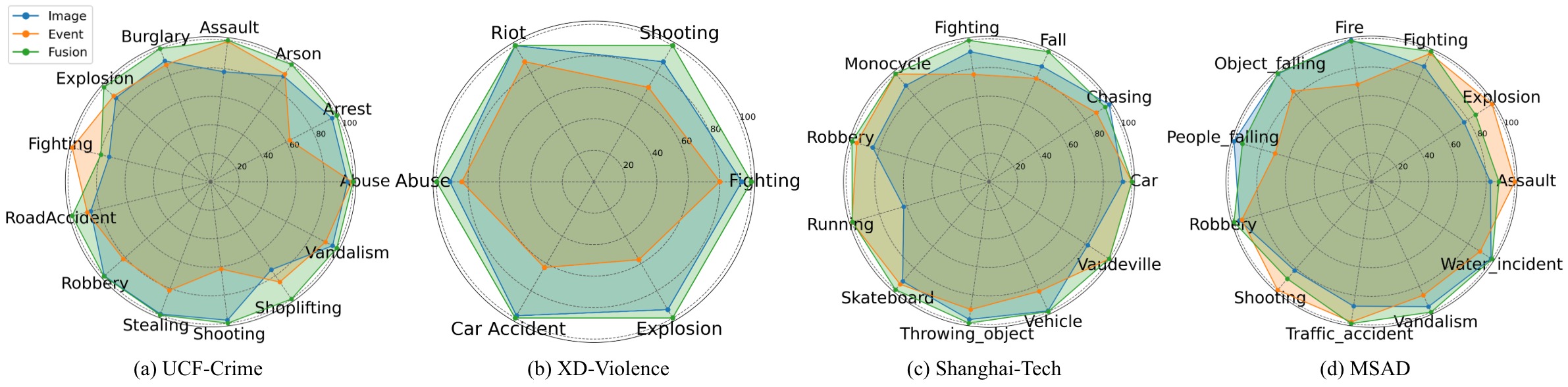}
    \caption{
        Radar charts showing per-class anomaly detection performance (AUC and AP) for image-only (\textcolor{blue}{blue}), motion-cue-only (\textcolor{orange}{orange}), and fused (\textcolor{mygreen}{green}) approaches. Each radial axis represents an anomaly category, and values are normalized per class by the maximum score. The fused approach consistently covers a larger area, highlighting improved detection across anomaly types.
    }
    \label{fig2}
\end{figure*}

        These trends are further illustrated in Fig.~\ref{fig2}, which provides a class-wise comparison of image-only, motion-cue-only, and fused approaches. In most anomaly classes, the image-based modality achieves higher performance than the motion-cue-based modality, reflecting the substantial differences in the underlying information each modality encodes. However, in anomaly categories that align more closely with the characteristics of $z_e$ (e.g., Fighting and Assault in UCF-Crime), the motion-cue-based modality outperforms the image-based one, confirming our initial intuition. By combining $z_x$ and $z_e$, our method consistently achieves higher detection accuracy than either single modality alone. The fusion captures the strengths of the image modality while incorporating the advantages of the motion-cue modality in the classes where it excels, leading to additional gains even in categories that are already well handled by image features. Consequently, as shown in Fig.~\ref{fig2}, the fused approach spans a broader region in the radial plots, surpassing the performance of single-modality baselines across most anomaly categories.
        
    \subsection{Fusion: Uncertainty Behavior Under Value-Level Masking}    
        \begin{figure*}[ht]
            \centering
            \includegraphics[width=1.0\linewidth]{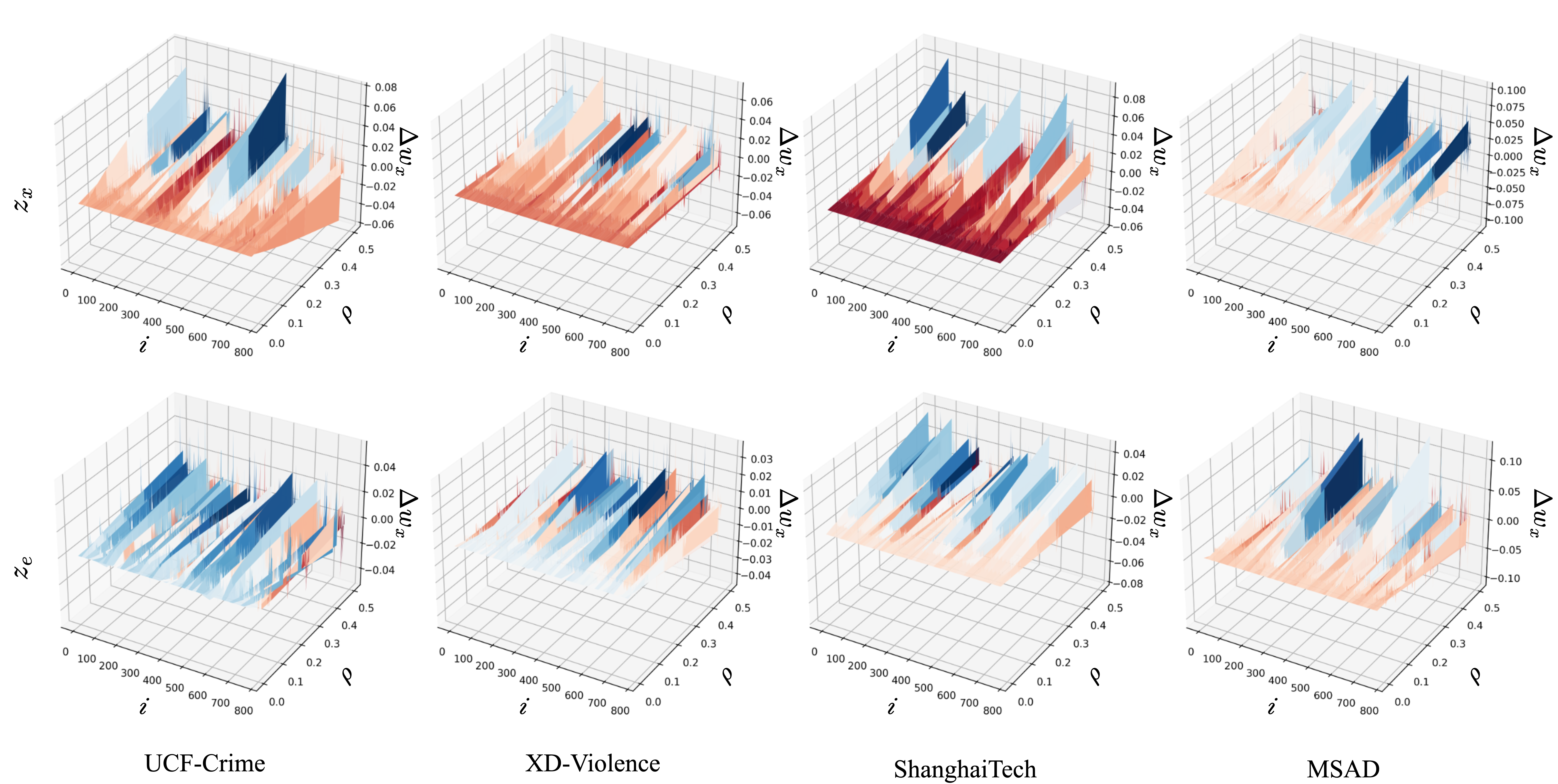}
            \caption{
                Change in $\Delta w_x$ under modality-specific masking perturbations. The horizontal axis $i$ is the latent dimension index, and the vertical axis $\rho$ the proportion of masked values in $z_x$ or $z_e$. Each surface shows $\Delta w_x(i,\rho) = w_x^{\text{masked}}(i) - w_x^{\text{clean}}(i)$ for a given masking ratio. The top row applies masking to $z_x$, the bottom to $z_e$, both measuring the change in $w_x$. Positive values (\textcolor{blue}{blue}) indicate increased image confidence under corruption, while negative values (\textcolor{red}{red}) indicate reduction. Non-uniform patterns across $i$ highlight dimension-specific responses to degradation.
            }
            \label{fig:uncertainty_change}
        \end{figure*}

\begin{table*}[ht]
\centering
\small
\setlength{\tabcolsep}{4pt}
\resizebox{\textwidth}{!}{
\begin{tabular}{lc|ccccc|ccccc|ccccc|cccccc}
\toprule
\multicolumn{2}{c|}{-} &
\multicolumn{5}{c|}{\textbf{UCF-Crime~\cite{sultani2018real} (AUC)}} &
\multicolumn{5}{c|}{\textbf{XD-Violence~\cite{wu2020not} (AP)}} &
\multicolumn{5}{c|}{\textbf{ShanghaiTech~\cite{liu2018future} (AUC)}} &
\multicolumn{5}{c}{\textbf{MSAD~\cite{msad2024} (AUC)}} \\
\midrule
\textbf{Noise} & \textbf{Mask} & \textbf{AUC} & \textbf{KL} 
& $\Delta w_x$ & $\Delta w_x^{ab}$ & $\Delta w_x^{n}$
& \textbf{AP} & \textbf{KL} 
& $\Delta w_x$ & $\Delta w_x^{ab}$ & $\Delta w_x^{n}$
& \textbf{AUC} & \textbf{KL} 
& $\Delta w_x$ & $\Delta w_x^{ab}$ & $\Delta w_x^{n}$
& \textbf{AUC} & \textbf{KL} 
& $\Delta w_x$ & $\Delta w_x^{ab}$ & $\Delta w_x^{n}$ \\
\midrule

CLEAN & 0 
& 89.09 & 0.0000 & 0.0000 & 0.0000 & 0.0000
& 88.26 & 0.0000 & 0.0000 & 0.0000 & 0.0000
& 98.17 & 0.0000 & 0.0000 & 0.0000 & 0.0000
& 92.39 & 0.0000 & 0.0000 & 0.0000 & 0.0000 \\
\midrule

\multirow{5}{*}{EV\_NOISE}
& 0.05 
& 88.92 & 0.0032 & +0.0003 & +0.0002 & +0.0003
& 88.12 & 0.0020 & 0.0000 & 0.0000 & 0.0000
& 98.13 & 0.0006 & -0.0002 & -0.0001 & -0.0002
& 92.27 & 0.0024 & -0.0001 & -0.0001 & -0.0001 \\

& 0.10 
& 88.80 & 0.0062 & +0.0004 & +0.0002 & +0.0003
& 88.00 & 0.0040 & -0.0001 & -0.0001 & -0.0000
& 98.11 & 0.0022 & -0.0004 & -0.0002 & -0.0004
& 92.19 & 0.0048 & -0.0003 & -0.0002 & -0.0003 \\

& 0.20 
& 88.68 & 0.0104 & +0.0003 & +0.0001 & +0.0002
& 87.74 & 0.0083 & -0.0001 & -0.0001 & -0.0001
& 98.07 & 0.0038 & -0.0008 & -0.0005 & -0.0007
& 91.94 & 0.0088 & -0.0005 & -0.0003 & -0.0006 \\

& 0.30 
& 88.64 & 0.0133 & +0.0003 & +0.0001 & +0.0002
& 87.51 & 0.0121 & -0.0002 & -0.0001 & -0.0002
& 98.06 & 0.0053 & -0.0011 & -0.0006 & -0.0010
& 91.89 & 0.0116 & -0.0007 & -0.0005 & -0.0007 \\

& 0.50 
& 88.35 & 0.0198 & +0.0002 & -0.0001 & +0.0001
& 87.13 & 0.0194 & -0.0007 & -0.0005 & -0.0006
& 98.06 & 0.0082 & -0.0016 & -0.0008 & -0.0015
& 91.86 & 0.0158 & -0.0010 & -0.0008 & -0.0009 \\
\midrule

\multirow{5}{*}{IMG\_NOISE}
& 0.05 
& 88.57 & 0.0258 & -0.0004 & -0.0005 & -0.0004
& 87.95 & 0.0267 & -0.0004 & -0.0005 & -0.0004
& 97.66 & 0.0162 & -0.0008 & -0.0014 & -0.0006
& 92.27 & 0.0066 & -0.0004 & -0.0003 & -0.0005 \\

& 0.10 
& 87.90 & 0.0548 & -0.0009 & -0.0011 & -0.0008
& 87.64 & 0.0546 & -0.0009 & -0.0010 & -0.0008
& 97.54 & 0.0308 & -0.0015 & -0.0016 & -0.0014
& 92.10 & 0.0125 & -0.0009 & -0.0008 & -0.0009 \\

& 0.20 
& 87.05 & 0.1066 & -0.0019 & -0.0022 & -0.0018
& 86.60 & 0.1162 & -0.0018 & -0.0022 & -0.0017
& 96.29 & 0.0758 & -0.0024 & -0.0034 & -0.0020
& 91.86 & 0.0220 & -0.0017 & -0.0010 & -0.0018 \\

& 0.30 
& 86.57 & 0.1445 & -0.0028 & -0.0032 & -0.0026
& 85.86 & 0.1788 & -0.0026 & -0.0030 & -0.0025
& 95.09 & 0.1084 & -0.0036 & -0.0045 & -0.0033
& 91.63 & 0.0303 & -0.0025 & -0.0021 & -0.0025 \\

& 0.50 
& 85.74 & 0.2021 & -0.0049 & -0.0054 & -0.0047
& 84.23 & 0.3082 & -0.0044 & -0.0049 & -0.0042
& 90.72 & 0.2180 & -0.0090 & -0.0105 & -0.0087
& 91.52 & 0.0425 & -0.0041 & -0.0038 & -0.0041 \\
\bottomrule
\end{tabular}}
\caption{
Unified comparison of fusion robustness under value-level corruption across four benchmarks.
Noise/Mask levels follow UCF-Crime; other datasets share the same degradation setting.
Image masking consistently reduces performance and $w_x$, while event masking yields minimal shifts across all datasets.}
\label{tab:four_dataset_fusion_comparison}
\end{table*}

        To investigate how the uncertainty weights $w_m$ respond to modality‑specific degradation, we perturb a single modality by masking a random subset of its latent features $z_m\in\mathbb R^{B\times T\times D}$.  A masked sequence is defined as
            \[
              z_{m,i}^{\text{masked}} =
              \begin{cases}
                0,          & i \in \mathcal I_\rho \\[4pt]
                z_{m,i},    & \text{otherwise}
              \end{cases},
              \qquad
              |\mathcal I_\rho| = \rho D .
            \]
        where $\rho\in(0,1)$ denotes the masking ratio and $i$ indexes feature dimensions.  We then measure the change in the image‑side uncertainty weight, $\Delta w_x = w_x^{\text{masked}}-w_x^{\text{clean}}$, with the normalisation $w_x+w_e=1$ enforced.  We prefer masking over additive noise because it produces a deterministic, localised degradation that preserves the Student‑$t$ noise assumption and avoids the non‑linear propagation artifacts that Gaussian perturbations introduce in attention layers, yielding a cleaner causal probe of the learned uncertainty mechanism.

        Fig.~\ref{fig:uncertainty_change} plots $\Delta w_x$ for three masking ratios ($\rho\in{0.05,0.10,0.20}$) on four datasets.  The top row masks $z_x$, consistently decreasing $w_x$—indicating higher image uncertainty—whereas the bottom row masks $z_e$, symmetrically increasing $w_x$.  The magnitude of $|\Delta w_x|$ grows with $\rho$, showing that the fusion network scales its confidence shift with corruption severity.
        
        Value‑level curves expose pronounced heterogeneity: some dimensions (e.g., indices 12–27) react strongly, whereas others remain flat.  We attribute this to distributed encoding: dimensions dominated by stable appearance cues are robust, while those capturing transient, modality‑specific dynamics are fragile.  The non‑uniform yet directionally consistent responses provide empirical evidence that the model estimates uncertainty on a fine‑grained basis rather than collapsing it into a scalar.

\begin{table*}[ht!]
\centering
\small
\renewcommand{\arraystretch}{1.08}
\setlength{\tabcolsep}{8pt}
\resizebox{\textwidth}{!}{
\begin{tabular}{lccccc}
\toprule
\textbf{Dataset} & \textbf{$\nu$=2} & \textbf{$\nu$=4} & \textbf{$\nu$=6} & \textbf{$\nu$=8} & \textbf{$\nu$=10} \\
\midrule
UCF-Crime~\cite{sultani2018real}      & 87.98$\pm$0.33 & 88.34$\pm$0.30 & 88.38$\pm$0.40 & 88.67$\pm$0.45 & 88.11$\pm$0.47 \\
XD-Violence~\cite{wu2020not}         & 87.31$\pm$0.77 & 87.39$\pm$0.65 & 87.03$\pm$0.52 & 87.63$\pm$0.54 & 86.64$\pm$0.51 \\
ShanghaiTech~\cite{liu2018future}    & 97.91$\pm$0.10 & 97.91$\pm$0.10 & 97.90$\pm$0.10 & 97.98$\pm$0.07 & 97.92$\pm$0.06 \\
MSAD~\cite{msad2024}                 & 91.78$\pm$0.34 & 91.79$\pm$0.29 & 91.84$\pm$0.42 & 92.90$\pm$0.27 & 91.64$\pm$0.27 \\
\midrule

\textbf{Dataset} & \textbf{$N$=0} & \textbf{$N$=10} & \textbf{$N$=20} & \textbf{$N$=30} & \textbf{$N$=40} \\
\midrule
UCF-Crime~\cite{sultani2018real}      & 86.77$\pm$0.18 & 88.67$\pm$0.45 & 88.41$\pm$0.43 & 88.48$\pm$0.28 & 88.60$\pm$0.31 \\
XD-Violence~\cite{wu2020not}         & 86.01$\pm$1.10 & 87.63$\pm$0.54 & 87.52$\pm$0.40 & 87.71$\pm$0.70 & 87.54$\pm$0.60 \\
ShanghaiTech~\cite{liu2018future}    & 97.94$\pm$0.08 & 97.98$\pm$0.07 & 97.90$\pm$0.12 & 97.90$\pm$0.11 & 97.93$\pm$0.07 \\
MSAD~\cite{msad2024}                 & 90.69$\pm$0.35 & 92.90$\pm$0.27 & 92.29$\pm$0.30 & 92.72$\pm$0.29 & 92.69$\pm$0.21 \\
\midrule

\textbf{Dataset} & \textbf{$\epsilon$=1e-4} & \textbf{--} & \textbf{$\epsilon$=1e-6} & \textbf{--} & \textbf{$\epsilon$=1e-8} \\
\midrule
UCF-Crime~\cite{sultani2018real}      & 88.46$\pm$0.42 & -- & 88.50$\pm$0.38 & -- & 88.67$\pm$0.45 \\
XD-Violence~\cite{wu2020not}         & 87.27$\pm$0.76 & -- & 87.44$\pm$0.79 & -- & 87.63$\pm$0.54 \\
ShanghaiTech~\cite{liu2018future}    & 97.91$\pm$0.12 & -- & 97.90$\pm$0.08 & -- & 97.98$\pm$0.07 \\
MSAD~\cite{msad2024}                 & 91.77$\pm$0.27 & -- & 91.81$\pm$0.33 & -- & 92.90$\pm$0.27 \\
\midrule

\textbf{Dataset} & \textbf{$\lambda_r$=0.1} & \textbf{$\lambda_r$=0.3} & \textbf{$\lambda_r$=0.5} & \textbf{$\lambda_r$=0.7} & \textbf{$\lambda_r$=0.9} \\
\midrule
UCF-Crime~\cite{sultani2018real}      & 87.92$\pm$0.38 & 88.13$\pm$0.24 & 88.67$\pm$0.45 & 88.21$\pm$0.34 & 87.91$\pm$0.34 \\
XD-Violence~\cite{wu2020not}         & 86.89$\pm$0.80 & 86.74$\pm$0.48 & 87.63$\pm$0.54 & 87.19$\pm$0.67 & 87.23$\pm$1.13 \\
ShanghaiTech~\cite{liu2018future}    & 97.86$\pm$0.07 & 97.90$\pm$0.11 & 97.98$\pm$0.07 & 97.90$\pm$0.11 & 97.92$\pm$0.13 \\
MSAD~\cite{msad2024}                 & 90.73$\pm$0.74 & 91.41$\pm$0.53 & 92.90$\pm$0.27 & 92.12$\pm$0.29 & 92.28$\pm$0.44 \\
\bottomrule
\end{tabular}
}
\caption{AUC under different combinations of $\nu$, $N$, $\epsilon$, and $\lambda_r$ across four datasets.}
\label{tab:merged_ablation_compact}
\end{table*}

        To complement the value-level analysis in Fig.~\ref{fig:uncertainty_change}, we further report dataset-level quantitative shifts under the same masking protocol (Table~\ref{tab:four_dataset_fusion_comparison}). Across all four datasets, event-side corruption produces almost no change in AUC/AP, KL divergence, or uncertainty weights, indicating that the fusion module does not over-rely on the motion-cue stream. In contrast, masking the image modality yields consistent decreases in AUC/AP accompanied by sharp increases in KL divergence and a monotonic drop in $w_x$, confirming that the model correctly identifies degraded appearance features and reallocates confidence toward the complementary event representation. The magnitude of this reallocation varies with dataset characteristics—UCF-Crime and XD-Violence show the strongest shifts due to their appearance dependence, whereas ShanghaiTech and MSAD remain stable because both modalities provide redundant cues. Together, these results validate that the proposed fusion mechanism learns a directionally consistent and dataset-aware uncertainty response rather than collapsing to a single dominant modality.

    \subsection{Ablation Study}
        \subsubsection{Hyperparameter Sensitivity}
            Table~\ref{tab:merged_ablation_compact} shows that all four hyperparameters exhibit stable and monotonic trends across datasets. The degrees of freedom $\nu$ improves performance up to $\nu{=}8$, after which gains saturate or slightly drop. The refinement steps $N$ provide the largest impact, with a clear jump from $N{=}0$ to $N{=}10$ and minimal change afterward, indicating rapid convergence of the refinement module. The Laplace precision $\epsilon$ has negligible influence, with variations staying within narrow margins for every dataset. The refinement weight $\lambda_r$ performs best in the mid-range (0.5–0.7), while extreme values weaken fusion quality. Overall, the table confirms that IEF-VAD remains robust under wide hyperparameter variation, and optimal performance emerges from moderate heavy-tailed modeling, early refinement, and balanced regularization.

\begin{table}[t]
    \centering
    \small
    \setlength{\tabcolsep}{4pt}
    \renewcommand{\arraystretch}{1.15}
    \begin{tabularx}{\textwidth}{l c *{4}{>{\centering\arraybackslash}X}}
        \toprule
        Dataset & Metric
        & $\mathcal{L}_{\textbf{cls}}$
        & $\mathcal{L}_{\textbf{cls}}+\mathcal{L}_{\textbf{reg}}$
        & $\mathcal{L}_{\textbf{cls}}+\mathcal{L}_{\textbf{KL}}$
        & $\mathcal{L}_{\textbf{cls}}+\mathcal{L}_{\textbf{reg}}+\mathcal{L}_{\textbf{KL}}$ \\
        \midrule
        \multirow{2}{*}{UCF-Crime~\cite{sultani2018real}}
            & AUC     & 88.04$\pm$0.28 & 87.78$\pm$0.46 & 88.24$\pm$0.19 & 88.67$\pm$0.45 \\
            & Ano-AUC & 69.79$\pm$0.85 & 69.73$\pm$1.00 & 70.46$\pm$0.69 & 71.50$\pm$1.02 \\
        \midrule
        XD-Violence~\cite{wu2020not}
            & AP      & 87.40$\pm$0.72 & 87.45$\pm$0.49 & 87.46$\pm$0.64 & 87.63$\pm$0.54 \\
        \midrule
        ShanghaiTech~\cite{liu2018future}
            & AUC     & 97.97$\pm$0.07 & 97.91$\pm$0.09 & 97.85$\pm$0.14 & 97.98$\pm$0.07 \\
        \midrule
        MSAD~\cite{msad2024}
            & AUC     & 92.39$\pm$0.22 & 91.65$\pm$0.32 & 92.38$\pm$0.22 & 92.90$\pm$0.27 \\
        \bottomrule
    \end{tabularx}
    \caption{Performance comparison under different loss configurations across datasets.}
    \label{tab:loss_ablation}
\end{table}

        \subsubsection{Fine-Grained Refinement Steps}
            \label{subsec:n_finegrained}
            To show how performance evolves over the early refinement range, Table~\ref{tab:n_finegrained} reports AUC at $N \in \{0,2,4,6,8,10\}$ on all four benchmarks. The gains accumulate gradually and largely converge by $N{=}10$, complementing the coarse $N$ sweep in Table~\ref{tab:merged_ablation_compact}.

            \begin{table}[h]
                \centering
                \small
                \setlength{\tabcolsep}{6pt}
                \caption{AUC under fine-grained refinement steps $N \in \{0,2,4,6,8,10\}$ across four datasets.}
                \label{tab:n_finegrained}
                \begin{tabular}{lcccccc}
                    \toprule
                    \textbf{Dataset} & $N{=}0$ & $N{=}2$ & $N{=}4$ & $N{=}6$ & $N{=}8$ & $N{=}10$ \\
                    \midrule
                    UCF-Crime~\cite{sultani2018real}   & 86.77 & 87.95 & 87.40 & 87.65 & 87.81 & \textbf{88.67} \\
                    XD-Violence~\cite{wu2020not}       & 86.01 & 86.32 & 87.42 & 87.78 & \textbf{88.10} & 87.63 \\
                    ShanghaiTech~\cite{liu2018future}  & 97.94 & 97.96 & 97.88 & 97.94 & \textbf{98.01} & 97.97 \\
                    MSAD~\cite{msad2024}               & 90.69 & 90.69 & 91.05 & 91.74 & 92.34 & \textbf{92.90} \\
                    \bottomrule
                \end{tabular}
            \end{table}

        \subsubsection{Event Extraction Parameters}
            \label{subsec:taut_sweep}
            The synthetic motion cue depends on two extraction hyperparameters: the temporal window $T$ over which activations are accumulated and the threshold $\tau$ applied to inter-frame differences. To verify that our results are not an artifact of a particular setting, we \emph{retrain} IEF-VAD from scratch on UCF-Crime for every combination of $T\in\{8,16,32\}$ and $\tau\in\{5,10,20\}/255$, reporting both AUC and anomaly-focused AUC in Table~\ref{tab:retrained_taut}. Across the grid the model stays within roughly $0.7$\,pp for eight of the nine settings, and the trained default $(T{=}16,\tau{=}10)$ is among the strongest, confirming that the fusion mechanism rather than a finely tuned cue extractor drives performance.

            \begin{table}[h]
                \centering
                \small
                \setlength{\tabcolsep}{4pt}
                \caption{UCF-Crime retrained sweep over the temporal window $T$ and threshold $\tau$; each cell is trained from scratch. $\ddagger$ marks the default configuration, and the best value in each row is in bold.}
                \label{tab:retrained_taut}
                \resizebox{\linewidth}{!}{%
                \begin{tabular}{l|ccccccccc}
                \toprule
                $(T,\ \tau{\cdot}255)$
                & (8,5) & (8,10) & (8,20)
                & (16,5) & (16,10)$^\ddagger$ & (16,20)
                & (32,5) & (32,10) & (32,20) \\
                \midrule
                AUC      & 89.19 & 89.22 & 88.58 & \textbf{89.25} & 89.13 & 88.83 & 88.86 & 88.74 & 87.09 \\
                Ano-AUC  & 72.98 & \textbf{73.01} & 71.42 & 72.80 & 72.49 & 71.66 & 72.04 & 71.50 & 70.23 \\
                \bottomrule
                \end{tabular}}
            \end{table}

        \subsubsection{Loss Configuration}
            Table~\ref{tab:loss_ablation} shows that each loss term contributes differently to fusion stability. Adding the KL component consistently improves performance across all datasets, while the regularization term alone yields only marginal gains. The full combination achieves the highest or comparable scores everywhere, indicating that variance control (via $\mathcal{L}_{\textbf{KL}}$) is the primary driver of robustness, and that $\mathcal{L}_{\textbf{reg}}$ mainly provides small alignment benefits without interfering with classification quality.
\section{Limitations and Future Work.} 
    While our study advances uncertainty-guided fusion, several limitations remain. First, modeling each modality’s noise with a diagonal covariance ignores cross-feature correlations; lightweight structured or low-rank estimators could better capture value-level dependencies. Second, fixed regularization weights $\lambda_{1,2}$ and Student-t degrees-of-freedom $\nu$ reduce adaptability; learning them jointly or via meta-learning would allow dynamic adjustment to modality quality. Third, replacing simple averaging with uncertainty-conditioned attention, gating, or message passing may yield a more expressive fusion mechanism that exploits inter-modal disagreement for stronger anomaly detection and broader multimodal perception.
     
\section{Conclusion}
    \label{conclusion}
    \textit{IEF-VAD} shows that motion-centric event-like representations synthetically extracted from ordinary videos can be combined with RGB frames to extend video anomaly detection beyond the limits of frame based models while maintaining high practicality since no dedicated event cameras are required. The method sets new state of the art results on UCF-Crime, XD-Violence, ShanghaiTech, and MSAD and a value level masking analysis confirms that its weights adjust across latent dimensions in a manner that reflects consistent use of modality specific cues. These results indicate that synthetic motion cue data offers a promising direction for broader multimodal video understanding and provide evidence that it can support principled and effective fusion frameworks, since it can be extracted directly from any video source and therefore applied flexibly across diverse settings without specialized hardware.

\section*{Acknowledgements}
    This work was supported in part by the DARPA Young Faculty Award, the National Science Foundation (NSF) under Grants \#2127780, \#2319198, \#2321840, \#2312517, and \#2235472, the Semiconductor Research Corporation (SRC), the Office of Naval Research through the Young Investigator Program Award, and Grants \#N00014-21-1-2225 and \#N00014-24-1-2547, Army Research Office Grant \#W911NF2410360. Additionally, support was provided by the Air Force Office of Scientific Research under Award \#FA9550-22-1-0253.

%
%
\bibliographystyle{splncs04}
\bibliography{main}
\end{document}